\pdfoutput=1

\documentclass[11pt]{article}

\usepackage{emnlp2021}

\usepackage{times}
\usepackage{latexsym}

\usepackage[T1]{fontenc}

\usepackage[utf8]{inputenc}

\usepackage{microtype}

\usepackage{times}
\usepackage{latexsym}

\usepackage{xcolor}
\usepackage{graphicx}
\usepackage{tabularx}
\usepackage{booktabs}
\usepackage{xspace}
\usepackage{array}
\usepackage{subcaption}
\usepackage{enumitem}

\newcolumntype{L}[1]{>{\raggedright\let\newline\\\arraybackslash\hspace{0pt}}m{#1}}
\newcolumntype{C}[1]{>{\centering\let\newline\\\arraybackslash\hspace{0pt}}m{#1}}
\newcolumntype{R}[1]{>{\raggedleft\let\newline\\\arraybackslash\hspace{0pt}}m{#1}}

\usepackage{microtype}

\graphicspath{ {figures/} }

\newcommand{\ccnet}{CCNet\xspace}
\newcommand{\wikilarge}{WikiLarge\xspace}
\newcommand{\asset}{ASSET\xspace}
\newcommand{\newsela}{Newsela\xspace}
\newcommand{\alector}{ALECTOR\xspace}

\newcommand{\turkcorpus}{TurkCorpus\xspace}
\newcommand{\turkcorpusabbr}{TurkCor.\xspace}

\newcommand{\muss}{\textsc{MUSS}\xspace}
\newcommand{\bart}{\textsc{BART}\xspace}
\newcommand{\mbart}{\textsc{mBART}\xspace}
\newcommand{\access}{\mbox{\textsc{ACCESS}}\xspace}
\newcommand{\bartaccess}{\textsc{BART}+\textsc{ACCESS}\xspace}
\newcommand{\mbartaccess}{\textsc{mBART}+\textsc{ACCESS}\xspace}
\newcommand{\laser}{LASER\xspace}

\newcommand{\mined}{\textsc{Mined}\xspace}

\newcommand{\numem}[2]{$#1{\scriptstyle\pm #2}$}
\newcommand{\tnumem}[2]{$#1$&\numem{}{#2}}

\title{MUSS: Multilingual Unsupervised Sentence Simplification\\ by Mining Paraphrases}
\author{Louis Martin$^{1,2}$\quad Angela Fan$^{1,3}$\\ \large\textbf{\'Eric de la Clergerie$^2$\quad Antoine Bordes$^1$\quad Beno\^it Sagot$^2$}\\
  $^{1}$Facebook AI Research, Paris, France \\
  $^{2}$Inria, Paris, France \\
  $^{3}$LORIA, Nancy, France\\
  \texttt{\{louismartin, angelafan, abordes\}@fb.com}\\
  \texttt{\{eric.de\_la\_clergerie, benoit.sagot\}@inria.fr}}
\begin{document}
\maketitle
\begin{abstract}
Progress in sentence simplification has been hindered by a lack of labeled  parallel simplification data, particularly in languages other than English.
We introduce \muss, a Multilingual Unsupervised Sentence Simplification system that does not require labeled simplification data.
\muss uses a novel approach to sentence simplification that trains strong models using sentence-level paraphrase data  instead of proper simplification data.
These models leverage unsupervised pretraining and controllable generation mechanisms to flexibly adjust attributes such as length and lexical complexity at inference time.
We further present a method to mine such paraphrase data in any language from Common Crawl using semantic sentence embeddings, thus removing the need for labeled data.
We evaluate our approach on English, French, and Spanish simplification benchmarks and closely match or outperform the previous best supervised results, despite not using any labeled simplification data.
We push the state of the art further by incorporating labeled simplification data.
\end{abstract}
\section{Introduction}

Sentence simplification is the task of making a sentence easier to read and understand by reducing its lexical and syntactic complexity, while retaining most of its original meaning. Simplification has a variety of important societal applications, for example increasing accessibility for those with cognitive disabilities such as aphasia \cite{carroll1998practical}, dyslexia \cite{rello2013simplify}, and autism \cite{evans2014evaluation}, or for non-native speakers \cite{paetzold2016unsupervised}. %
Research has mostly focused on English simplification, where source texts and  associated simplified texts exist and can be automatically aligned, such as English Wikipedia and Simple English Wikipedia \cite{zhang2017sentence}. %
However, such data is limited in terms of size and domain, and difficult to find in other languages.
Additionally, simplifying a sentence can be achieved in multiple ways, and depend on the target audience. %
Simplification guidelines are not uniquely defined, outlined by the stark differences in English simplification benchmarks \cite{alva2020asset}.
This highlights the need for more general models that can adjust to different simplification contexts and scenarios.

In this paper, we propose to train controllable models using sentence-level paraphrase data only, i.e. parallel sentences that have the same meaning but phrased differently.
In order to generate simplifications and not paraphrases, we use \access \cite{martin2020controllable} to control attributes such as length, lexical and syntactic complexity.
Paraphrase data is more readily available, and opens the door to training flexible models that can adjust to more varied simplification scenarios.
We propose to gather such paraphrase data in any language by mining sentences from Common Crawl using semantic sentence embeddings.
We show that simplification models trained on mined paraphrase data perform as well as models trained on existing large paraphrase corpora (cf.~Appendix~\ref{section:comparison_with_paraphrase_dataset}). Moreover, paraphrases  are  more straightforward to mine than simplifications, and we show that they lead to models with better performance than equivalent models trained on mined simplifications (cf.~Section~\ref{subsection:ablations}).

Our resulting Multilingual Unsupervised Sentence Simplification method, \muss, is \textit{unsupervised} because it can be trained without relying on {\em labeled} simplification data,\footnote{We use the term {\em labeled simplifications} to refer to parallel datasets where texts were manually simplified by humans.} even though we mine using supervised sentence embeddings.\footnote{Previous works have also used the term {\em unsupervised simplification} to describe works that do not use any labeled parallel simplification data while leveraging supervised components such as constituency parsers and knowledge bases \citep{kumar-etal-2020-iterative}, external synonymy lexicons \cite{surya2018unsupervised}, and databases of simplified synonyms \cite{Zhao2020SemiSupervisedTS}. We shall come back to these works in Section~\ref{sec:related_work}.}
We additionally incorporate unsupervised pretraining  \cite{liu2019unsupervised,liu2020multilingual} and apply \muss on English, French, and Spanish to closely match or outperform the supervised state of the art in all languages. \muss further improves the state of the art on all English datasets by incorporating additional labeled simplification data.

To sum up, our contributions are as follows:
\begin{itemize}
    \item We introduce a novel approach to training simplification models with paraphrase data only and propose a mining procedure to create large paraphrase corpora for any language.
    \item Our approach obtains strong performance. Without any labeled simplification data, we match or outperform the supervised state of the art in English, French and Spanish. We further improve the English state of the art by incorporating labeled simplification data.
    \item We release \muss pretrained models, paraphrase data, and code for mining and training\footnote{\url{https://github.com/facebookresearch/muss}}.
\end{itemize}

\section{Related work}\label{sec:related_work}

Data-driven methods have been predominant in {\bf English sentence simplification} in  recent years \cite{alva2020data}, requiring large supervised training corpora of complex-simple aligned sentences \cite{wubben2012sentence,xu2016optimizing,zhang2017sentence,zhao2018integrating,martin2020controllable}.
Methods have relied on English and Simple English Wikipedia with automatic sentence alignment from similar articles \cite{zhu2010monolingual,coster2011learning,woodsend2011learning,kauchak2013improving,zhang2017sentence}. %
Higher quality datasets have been proposed such as the \newsela corpus \cite{xu2015problems}, but they are rare and come with restrictive licenses that hinder reproducibility and widespread usage.

{\bf Simplification in other languages} has been explored in Brazilian Portuguese \cite{aluisio2008towards}, Spanish \cite{saggion2015making,vstajner2015automatic}, Italian \cite{brunato2015design,tonelli2016simpitiki}, Japanese \cite{goto2015japanese,kajiwara2018text,katsuta2019improving}, and French \cite{gala2020alector}, but the lack of a large labeled parallel corpora has slowed research down.
In this work, we show that a method trained on automatically mined corpora can reach state-of-the-art results in each language.

When labeled parallel simplification data is unavailable, systems rely on {\bf unsupervised simplification} techniques, often inspired from machine translation.
The prevailing approach is to split a monolingual corpora into disjoint sets of complex and simple sentences using readability metrics. Then simplification models can be trained by using automatic sentence alignments \cite{kajiwara-komachi-2016-building,kajiwara2018text}, auto-encoders \cite{surya2018unsupervised,Zhao2020SemiSupervisedTS}, unsupervised statistical machine translation \cite{katsuta2019improving}, or back-translation \cite{aprosio2019neural}.
Other unsupervised simplification approaches iteratively edit the sentence until a certain simplicity criterion is reached \cite{kumar-etal-2020-iterative}.
The performance of unsupervised methods are often below their supervised counterparts.
\muss bridges the gap with supervised method and removes the need for deciding in advance how complex and simple sentences should be separated, but instead trains directly on paraphrases mined from the raw corpora.

Previous work on {\bf parallel dataset mining} have been used mostly in machine translation using document retrieval \cite{munteanu-marcu-2005-improving}, language models \cite{koehn-etal-2018-findings,koehn2019findings},
and embedding space alignment \cite{artetxe2019massively} to create large corpora \cite{tiedemann2012parallel,schwenk2019ccmatrix}.
We focus on paraphrasing for sentence simplifications, which presents new challenges. Unlike machine translation, where the same sentence should be identified in two languages, we develop a method to identify varied paraphrases of sentences, that have a wider array of surface forms, including different lengths, multiple sentences, different vocabulary usage, and removal of content from more complex sentences. 

Previous {\bf unsupervised paraphrasing} research has aligned sentences from various parallel corpora \cite{barzilay2003learning} with multiple objective functions \cite{liu2019unsupervised}.
Bilingual pivoting relied on MT datasets to create large databases of word-level paraphrases \cite{pavlick2015ppdb}, lexical simplifications \cite{pavlick2016simple,kriz-etal-2018-simplification}, or sentence-level paraphrase corpora \cite{wieting-gimpel-2018-paranmt}.
This has not been applied to multiple languages or to sentence-level simplification. Additionally, we use raw monolingual data to create our paraphrase corpora instead of relying on parallel MT datasets.

\section{Method}

In this section we describe \muss, our approach to mining paraphrase data and training controllable simplification models on paraphrases.

\begin{table}
\centering\small
\resizebox{\columnwidth}{!}{
\begin{tabular}{llrr}
\toprule
 & \multicolumn{1}{c}{\textbf{Type}} & \multicolumn{1}{c}{\textbf{\# Sequence}} & \multicolumn{1}{c}{\textbf{\# Avg. Tokens}} \\
 &  & \multicolumn{1}{c}{\textbf{Pairs}} & \multicolumn{1}{c}{\textbf{per Sequence}}\\ \midrule
 \textbf{\wikilarge} & Labeled Parallel & 296,402 & original: 21.7 \\
 \bf (English)& Simplifications&& simple: 16.0 \\
 \textbf{\newsela} & Labeled Parallel & 94,206 & original: 23.4 \\
 \bf (English)& Simplifications&& simple: 14.2 \\
 \midrule
\textbf{English} & Mined & 1,194,945 & 22.3 \\
\textbf{French} & Mined & 1,360,422 & 18.7 \\
\textbf{Spanish} & Mined & 996,609 & 22.8 \\
\bottomrule
\end{tabular}
}
\caption{\label{table:data_stats} Statistics on our mined paraphrase training corpora compared to standard simplification datasets (see section \ref{subsection:training_data} for more details).}
\end{table}

\subsection{Mining Paraphrases in Many Languages}

\paragraph{Sequence Extraction}
Simplification consists of multiple rewriting operations, some of which span  multiple sentences (e.g.~sentence splitting or fusion).
To allow such operations to be represented in our paraphrase data, we extract chunks of texts composed of multiple sentences, we refer to these small pieces of text by \emph{sequences}.

We extract such sequences by first tokenizing a document into individual sentences $\{s_1, s_2, \ldots, s_n\}$ using the NLTK sentence tokenizer \cite{loper2002nltk}.
We then extract sequences of adjacent sentences with maximum length of 300 characters: e.g. $\{[s_1], [s_1, s_2], [s_1, \ldots, s_k], [s_2], [s_2, s_3], ...\}$.
We can thus align two sequences that contain a different number of sentences, and represent sentence splitting or sentence fusion operations.

These sequences are further filtered to remove noisy sequences with more than 10\% punctuation characters and sequences with low language model probability  according to a 3-gram Kneser-Ney language model trained with \texttt{kenlm} \cite{heafield2011kenlm} on Wikipedia.

We extract these sequences from \ccnet \cite{wenzek2019ccnet}, an extraction of Common Crawl %
(an open source snapshot of the web) that has been split into different languages using \texttt{fasttext} language identification \cite{joulin2017bag} and various language modeling filtering techniques to identify high quality, clean sentences. For English and French, we extract 1 billion sequences from \ccnet. For Spanish we extract 650 millions sequences, the maximum for this language in \ccnet after filtering out noisy text.

\paragraph{Creating a Sequence Index Using Embeddings}
To automatically mine our paraphrase corpora, we first compute $n$-dimensional embeddings for each extracted sequence using \laser \cite{artetxe2019massively}. \laser provides joint multilingual sentence embeddings in 93 languages that have been successfully applied to the task of bilingual bitext mining \cite{schwenk2019ccmatrix}. In this work, we show that \laser can also be used to mine monolingual paraphrase datasets.

\paragraph{Mining Paraphrases}

After computing the embeddings for each language, we index them for fast nearest neighbor search using \texttt{faiss}. %

Each of these sequences is then used as a query $q_i$ against the billion-scale index that returns a set of top-8 nearest neighbor sequences according to the semantic \laser embedding space using L2 distance, resulting in a set of candidate paraphrases are $\{c_{i,1}, \ldots, c_{i,8}\}$.

We then use an upper bound on L2 distance and a margin criterion following \cite{artetxe-schwenk-2019-margin} to filter out sequences with low similarity.
We refer the reader to Appendix Section~\ref{sec:mining_details} for technical details.

The resulting nearest neighbors constitute a set of aligned paraphrases of the query sequence: $\{(q_i, c_{i, 1}), \ldots, (q_i, c_{i, j})\}$.
We finally apply poor alignment filters. We remove sequences that are almost identical with character-level Levenshtein distance $\leq$ 20\%, when they are contained in one another, or when they were extracted from two overlapping sliding windows of the same original document.

We report statistics of the mined corpora in English, French and Spanish in Table~\ref{table:data_stats}, and qualitative examples of the resulting mined paraphrases in Appendix Table~\ref{table:mining_examples}.
Models trained on the resulting mined paraphrases obtain similar performance than models trained on existing paraphrase datasets (cf. Appendix Section~\ref{section:comparison_with_paraphrase_dataset}).

\subsection{Simplifying with \access} \label{subsection:access}
In this section we describe how we adapt \access \cite{martin2020controllable} to train controllable models on mined paraphrases, instead of labeled parallel simplifications.

\access is a method to make any seq2seq model controllable by conditioning on simplification-specific control tokens.
We apply it on our seq2seq pretrained transformer models based on the \bart \cite{lewis2019bart} architecture (see next subsection).

\paragraph{Training with Control Tokens} At training time, the model is provided with control tokens that give oracle information on the target sequence, such as the amount of compression of the target sequence relative to the source sequence (length control). For example, when the target sequence is 80\% of the length of the original sequence, we provide the $<$NumChars\_80\%$>$ control token. At inference time we can then control the generation by selecting a given target control value.
We adapt the original Levenshtein similarity control to only consider replace operations but otherwise use the same controls as \citet{martin2020controllable}.
The controls used are therefore character length ratio, \textit{replace-only} Levenshtein similarity, aggregated word frequency ratio, and dependency tree depth ratio.
For instance we will prepend to every source in the training set the following 4 control tokens with sample-specific values, so that the model learns to rely on them: $<$NumChars\_XX\%$>$
$<$LevSim\_YY\%$>$ $<$WordFreq\_ZZ\%$>$ $<$DepTreeDepth\_TT\%$>$.
We refer the reader to the original paper \cite{martin2020controllable} and Appendix~\ref{subsection:training_details} for details on \access and how those control tokens are computed.

\paragraph{Selecting Control Values at Inference}
Once the model has been trained with oracle controls, we can adjust the control tokens to obtain the desired type of simplifications.
Indeed, sentence simplification often depends on the context and target audience \cite{martin2020controllable}.
For instance shorter sentences are more adapted to people with cognitive disabilities, while using more frequent words are useful to second language learners.
It is therefore important that supervised and unsupervised simplification systems can be adapted to different conditions: \citep{kumar-etal-2020-iterative} do so by choosing a set of operation-specific weights of their unsupervised simplification model for each evaluation dataset, \citep{surya2018unsupervised} select different models using SARI on each validation set.
Similarly, we set the 4 control hyper-parameters of \access using SARI on each validation set and keep them fixed for all samples in the test set.\footnote{Details in Appendix~\ref{subsection:training_details}}.
These 4 control hyper-parameters are intuitive and easy to interpret: when no validation set is available, they can also be set using prior knowledge on the task and still lead to solid performance (cf. Appendix~\ref{section:prior_knowledge_values}).

\subsection{Leveraging Unsupervised Pretraining}
We combine our controllable models with unsupervised pretraining to further extend our approach to text simplification. For English, we finetune the pretrained generative model \bart \cite{lewis2019bart} on our newly created training corpora. \bart is a pretrained sequence-to-sequence model that can be seen as a generalization of other recent pretrained models such as BERT \cite{devlin2018bert}.
For non-English, we use its multilingual generalization \mbart \cite{liu2020multilingual}, which was pretrained on 25 languages.
{}
\section{Experimental Setting}
We assess the performance of our approach on three languages: English, French, and Spanish.
We detail our experimental procedure for mining and training in Appendix Section~\ref{sec:experimental_details}.
In all our experiments, we report scores on the test sets averaged over 5 random seeds with 95\% confidence intervals.

\begin{table*}[!htbp]
\centering\small
\begin{tabular}{l|r@{}lr@{}l|r@{}lr@{}l|r@{}lr@{}l}
\multicolumn{1}{c}{} & \multicolumn{12}{c}{\textbf{English}} \\
\toprule
 & \multicolumn{4}{c|}{\textbf{\asset}} & \multicolumn{4}{c|}{\textbf{\turkcorpus}} & \multicolumn{4}{c}{\textbf{\newsela}} \\

  & \multicolumn{2}{c}{SARI $\uparrow$} & \multicolumn{2}{c|}{FKGL $\downarrow$} & \multicolumn{2}{c}{SARI $\uparrow$} & \multicolumn{2}{c|}{FKGL $\downarrow$} & \multicolumn{2}{c}{SARI $\uparrow$} & \multicolumn{2}{c}{FKGL $\downarrow$}\\
\midrule

\multicolumn{2}{l}{}\\[-2mm]  %
\multicolumn{5}{l}{\textbf{\textit{Baselines and Gold Reference}}} \\
\midrule
Identity Baseline & ${20.73}$& & ${10.02}$& & ${26.29}$& & ${10.02}$& & ${12.24}$& & ${8.82}$ \\
Truncate Baseline & ${29.85}$& & ${7.91}$& & ${33.10}$& & ${7.91}$ & & ${25.49}$ & & ${6.68}$ \\
Gold Reference & \tnumem{44.87}{0.36} & \tnumem{6.49}{0.15} & \tnumem{40.04}{0.30} & \tnumem{8.77}{0.08} & ---& & ---& \\
\midrule

\multicolumn{2}{l}{}\\[-2mm]  %
\multicolumn{2}{l}{\textbf{\textit{Unsupervised Systems}}} \\
\midrule
BTRLTS {\tiny \cite{Zhao2020SemiSupervisedTS}} & $33.95$ & & $7.59$ & & $33.09$ & & $8.39$ & & $37.22$ & & $3.80$ \\
UNTS {\tiny \cite{surya2018unsupervised}} & $35.19$& & $7.60$& & $36.29$& & $7.60$& & ---& & ---& \\
RM+EX+LS+RO {\tiny \cite{kumar-etal-2020-iterative}} & $36.67$ & & $7.33$ & & $37.27$ & & $7.33$ & & $\mathbf{38.33}$ & & $2.98$ \\
\midrule
\muss & \tnumem{\mathbf{42.65}}{0.23} & \tnumem{8.23}{0.62} & \tnumem{\mathbf{40.85}}{0.15} & \tnumem{8.79}{0.30} & \tnumem{\mathbf{38.09}}{0.59} & \tnumem{5.12}{0.47} \\
\midrule

\multicolumn{2}{l}{}\\[-2mm]  %
\multicolumn{2}{l}{\textbf{\textit{Supervised Systems}}} \\
\midrule
EditNTS {\tiny \cite{dong2019editnts}} & $34.95$ & & $8.38$ & & $37.66$ & & $8.38$ & & ${39.30}$& & ${3.90}$& \\
Dress-LS {\tiny \cite{zhang2017sentence}} & $36.59$& & $7.66$& & $36.97$& & $7.66$& & ${38.00}$& & ${4.90}$\\
DMASS-DCSS {\tiny \cite{zhao2018integrating}} & $38.67$& & $7.73$& & $39.92$& & $7.73$& & --- & & ---\\ 
\access {\tiny \cite{martin2020controllable}} & $40.13$& & $7.29$& & $41.38$& & $7.29$& & ---& & ---& \\ 
\midrule
\muss & \tnumem{\mathbf{43.63}}{0.71} & \tnumem{6.25}{0.42} & \tnumem{\mathbf{42.62}}{0.27} & \tnumem{6.98}{0.95} & \tnumem{\mathbf{42.59}}{1.00} & \tnumem{2.74}{0.98} \\
\muss (+ mined data) & \tnumem{\mathbf{44.15}}{0.56} & \tnumem{6.05}{0.51} & \tnumem{\mathbf{42.53}}{0.36} & \tnumem{7.60}{1.06} & \tnumem{\mathbf{41.17}}{0.95} & \tnumem{2.70}{1.00} \\

\bottomrule
\end{tabular}

\caption{\label{table:english_results} \textbf{Unsupervised and Supervised Sentence Simplification for English.} We display SARI and FKGL on \asset, \turkcorpus and \newsela test sets for English. Supervised models are trained on \wikilarge for the first two test sets, and \newsela for the last. Best SARI scores within confidence intervals are in bold.
}
\end{table*}
\subsection{Baselines}

In addition to comparisons with previous works, we implement and hereafter describe multiple baselines to assess the performance of our models, especially for French and Spanish where no previous simplification systems are available.

\paragraph{Identity} The entire original sequence is kept unchanged and used as the simplification. 

\paragraph{Truncation} The original sequence is truncated to the first 80\% words. It is a strong baseline according to standard simplification metrics.

\paragraph{Pivot} We use machine translation to provide a baseline for languages for which no simplification corpus is available. The source non-English sentence is translated to English, simplified with our best supervised English simplification system, and then translated back into the source language. For French and Spanish translation, we use \textsc{ccmatrix} \cite{schwenk2019ccmatrix} to train Transformer models with LayerDrop \cite{fan2019reducing}. We use the \bartaccess supervised model trained on \mined + \wikilarge as the English simplification model.
While pivoting creates potential errors, recent improvements of translation systems on high resource languages make this a strong baseline.

\paragraph{Gold Reference} We report gold reference scores for \asset and \turkcorpus as multiple references are available. We compute scores in a leave-one-out scenario where each reference is evaluated against all others. The scores are then averaged over all references.

\subsection{Evaluation Metrics}

We evaluate with the standard metrics SARI and FKGL. We report BLEU \cite{papineni2002bleu} only in Appendix Table~\ref{table:full_english_results} due its dubious suitability for sentence simplification \cite{sulem2018semantic}.
\paragraph{SARI} Sentence simplification is commonly evaluated with SARI \cite{xu2016optimizing}, which compares model-generated simplifications with the source sequence and gold references. It averages F1 scores for addition, keep, and deletion operations. We compute SARI with the  \texttt{EASSE} %
simplification evaluation suite \cite{alva2019easse}.\footnote{We use the latest version of SARI implemented in \texttt{EASSE} \cite{alva2019easse} which fixes bugs and inconsistencies from the traditional implementation. We thus recompute scores from previous systems that we compare to, by using the system predictions provided by the respective authors available in \texttt{EASSE}.}

\paragraph{FKGL} We report readability scores using the Flesch-Kincaid Grade Level (FKGL) \cite{kincaid1975derivation}, a linear combination of sentence lengths and word lengths.
FKGL was designed to be used on English texts only, we do not report it on French and Spanish.

\begin{table}
\centering\small
\begin{tabular}{l|r@{}l|r@{}l}
\toprule
 & \multicolumn{2}{c|}{\textbf{French}} & \multicolumn{2}{c}{\textbf{Spanish}} \\
 \textbf{\textit{{Baselines}}}  & \multicolumn{2}{C{2cm}|}{SARI $\uparrow$}& \multicolumn{2}{C{2cm}}{SARI $\uparrow$}\\
\midrule
Identity & ~~~~~~${26.16}$& &${16.99}$& \\
Truncate & ${33.44}$& & ~~~~~~${27.34}$& \\
Pivot & \tnumem{33.48}{0.37} & \tnumem{\mathbf{36.19}}{0.34} \\
\midrule
\textsc{MUSS}$\dagger$ & \tnumem{\mathbf{41.73}}{0.67} & \tnumem{\mathbf{35.67}}{0.46} \\
\bottomrule
\end{tabular}
\caption{\label{table:french_and_spanish_results} \textbf{Unsupervised Sentence Simplification in French and Spanish.} We display SARI scores in French (\alector) and  Spanish (\newsela). Best SARI scores within confidence intervals are in bold. $\dagger$\mbartaccess model.}
\end{table}

\subsection{Training Data} \label{subsection:training_data}
For all languages we use the mined data described in Table~\ref{table:data_stats} as training data.
We show that training with additional labeled simplification data leads to even better performance for English.
We use the labeled datasets \textbf{\wikilarge} \cite{zhang2017sentence} and \textbf{\newsela} \cite{xu2015problems}.
\wikilarge is composed of 296k simplification pairs automatically aligned from English Wikipedia and Simple English Wikipedia.
\newsela is a collection of news articles with professional simplifications, aligned into 94k simplification pairs by \citet{zhang2017sentence}.\footnote{We experimented with other alignments (wiki-auto and newsela-auto \cite{jiang2020neural}) but with lower performance.}

\subsection{Evaluation Data}\label{subsection:evaluation_data}

\paragraph{English} We evaluate our English models on \textbf{\asset} \cite{alva2020asset}, \textbf{\turkcorpus} \cite{xu2016optimizing} and \textbf{\newsela} \cite{xu2015problems}.
\turkcorpus  and \asset were created using the same 2000 valid and 359 test source sentences.
\turkcorpus contains 8 reference simplifications per source sentence and \asset contains 10 references per source.
\asset is a generalization of \turkcorpus with a more varied set of rewriting operations, and considered simpler by human judges \cite{alva2020asset}.
For \newsela, we evaluate on the split from \cite{zhang2017sentence}, which includes 1129 validation and 1077 test sentence pairs.

\paragraph{French} For French, we use the \textbf{\alector} dataset \cite{gala2020alector} for evaluation. \alector is a collection of literary (tales, stories) and scientific (documentary) texts along with their manual document-level simplified versions. These documents were extracted from material available to French primary school pupils. We split the dataset in 450 validation and 416 test sentence pairs (see Appendix~\ref{subsection:evaluation_details} for details).

\paragraph{Spanish} For Spanish we use the \textbf{Spanish part of \newsela} \cite{xu2015problems}.
We use the alignments from \cite{aprosio2019neural}, composed of 2794 validation and 2795 test sentence pairs.
Even though sentences were aligned using the \textsc{CATS} simplification alignment tool \cite{stajner-etal-2018-cats}, some alignment errors remain and automatic scores should be taken with a pinch of salt.

\section{Results}

\subsection{English Simplification}

We compare models trained on our mined corpus (see Table~\ref{table:data_stats}) with models trained on labeled simplification datasets (\wikilarge and \newsela). We also compare to other state-of-the-art supervised models: %
Dress-LS \cite{zhang2017sentence}, DMASS-DCSS \cite{zhao2018integrating}, EditNTS \cite{dong2019editnts}, \access \cite{martin2020controllable}; and the unsupervised models: UNTS \cite{surya2018unsupervised}, BTRLTS \cite{Zhao2020SemiSupervisedTS}, and RM+EX+LS+RO \cite{kumar-etal-2020-iterative}.
Results are shown in Table~\ref{table:english_results}.

\paragraph{\muss Unsupervised Results}
On the \asset benchmark, with no labeled simplification data, \muss obtains a +5.98 SARI improvement with respect to previous unsupervised methods, and a +2.52 SARI improvement over the state-of-the-art supervised methods.
For the \turkcorpus and \newsela datasets, the unsupervised \muss approach achieves strong results, either outperforming or closely matching both unsupervised and supervised previous works.

When incorporating labeled data from \wikilarge and \newsela, \muss obtains state-of-the-art results on all datasets.
Using labeled data along with mined data does not always help compared to training only with labeled data, especially with the \newsela training set. \newsela is already a high quality dataset focused on the specific domain of news articles. It might not benefit from additional lesser quality mined data.

\paragraph{Examples of Simplifications}
\begin{table*}[t]
\centering
\small
\begin{tabular}{@{}lp{0.87\textwidth}@{}}
\toprule

\textbf{Original} & \textbf{History} Landsberg prison, which is \textbf{in} the \textbf{town's} \textbf{western} \textbf{outskirts,} was \textbf{completed} in 1910. \\
\textbf{Simplified} & \textbf{The} Landsberg prison, which is \textbf{near} the \textbf{town,} was \textbf{built} in 1910. \\
\midrule

\textbf{Original} & The name "hornet" is used for this and related species \textbf{primarily} because \textbf{of} \textbf{their} \textbf{habit} \textbf{of} \textbf{making} \textbf{aerial} nests \textbf{(similar} \textbf{to} the true hornets) rather than \textbf{subterranean} \textbf{nests.} \\
\textbf{Simplified} & The name "hornet" is used for this and related species because \textbf{they} \textbf{make} nests \textbf{in} the \textbf{air} \textbf{(like} \textbf{the} true hornets) rather than \textbf{in} \textbf{the} \textbf{ground.} \\
\midrule
\textbf{Original} & Nocturnes is \textbf{an} \textbf{orchestral} \textbf{composition} \textbf{in} \textbf{three} \textbf{movements} by the French composer Claude Debussy. \\
\textbf{Simplified} & Nocturnes is \textbf{a} \textbf{piece} \textbf{of} \textbf{music} \textbf{for} \textbf{orchestra} by the French composer Claude Debussy. \\

\bottomrule
\end{tabular}
\caption{\textbf{Examples of Generated Simplifications.} We show simplifications generated by our best unsupervised model: \muss trained on mined data only. Bold highlights differences between original and simplified.}
\label{table:simplification_examples}
\end{table*}
Various examples from our unsupervised system are shown in Table~\ref{table:simplification_examples}. Examining the simplifications, we see reduced sentence length, sentence splitting, and simpler vocabulary usage. For example, the words \textit{in the town's western outskirts} is changed into \textit{near the town} and \textit{aerial nests} is simplified into \textit{nests in the air}.
We also witnessed errors related factual consistency and especially with respect with named entity hallucination or disappearance which would be an interesting area of improvement for future work.

\subsection{French and Spanish Simplification}

Our unsupervised approach to simplification can be applied to any language.
Similar to English, we first create a corpus of paraphrases composed of 1.4 million sequence pairs in French and 1.0 million sequence pairs in Spanish (cf. Table~\ref{table:data_stats}).
To incorporate multilingual pretraining, we replace the monolingual \bart with  \mbart, which was trained on 25 languages.

We report the performance of models trained on the mined corpus in Table~\ref{table:french_and_spanish_results}. Unlike English, where labeled parallel training data has been created using Simple English Wikipedia, no such datasets exist for French or Spanish. Similarly, no other simplification systems are available in these languages. We thus compare to several baselines, namely the identity, truncation and the strong pivot baseline.

\begin{figure*}[!htbp]
    \centering
\begin{subfigure}{0.30\textwidth}
    \centering
    \includegraphics[width=1\linewidth]{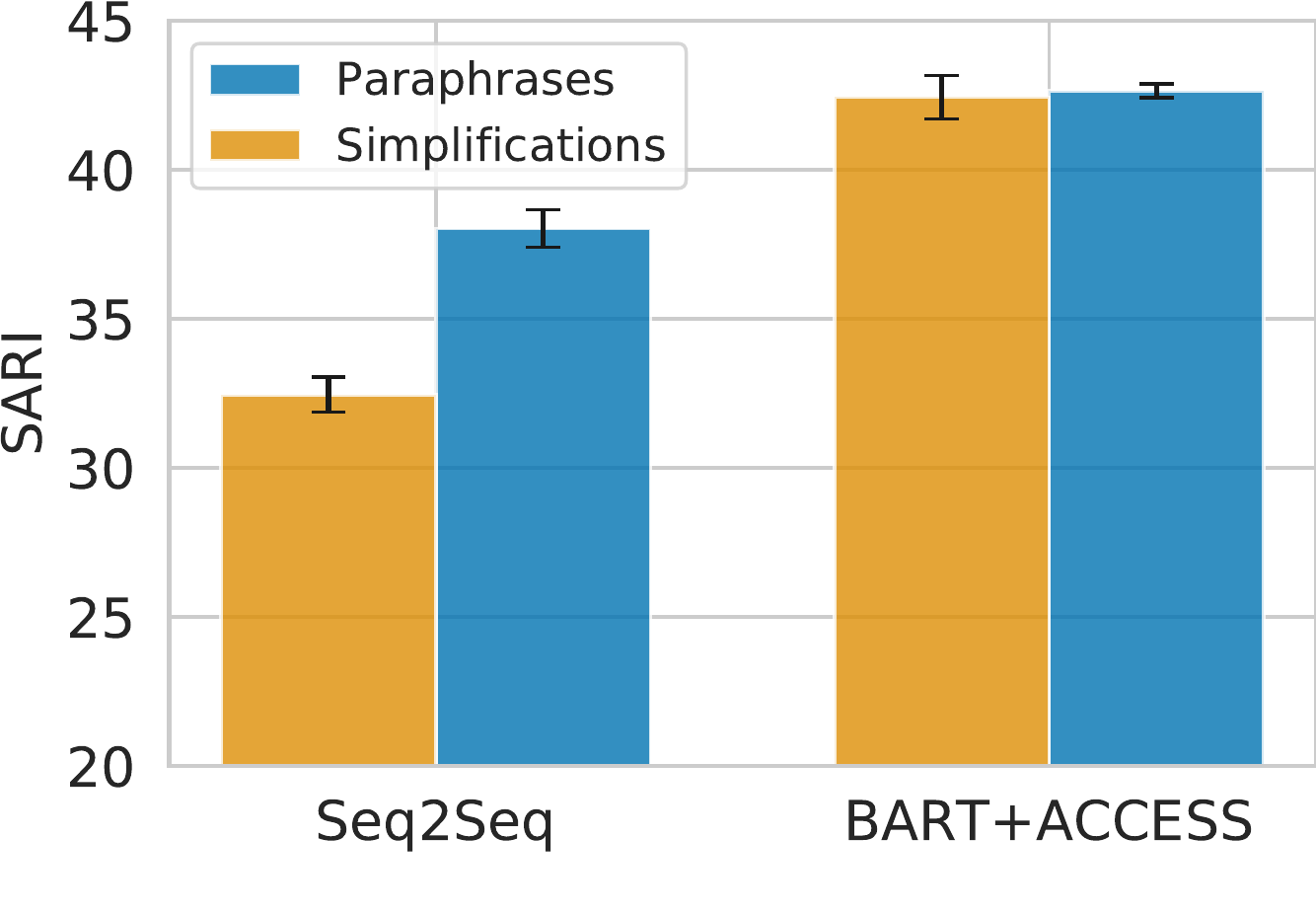}
    \caption{\label{subfigure:ablation_simplifications_vs_paraphrases} \textbf{Simplifications vs.~\mbox{Paraphrases}}}
\end{subfigure}
\hspace{1em}
\begin{subfigure}{0.30\textwidth}
    \centering
    \includegraphics[width=1\linewidth]{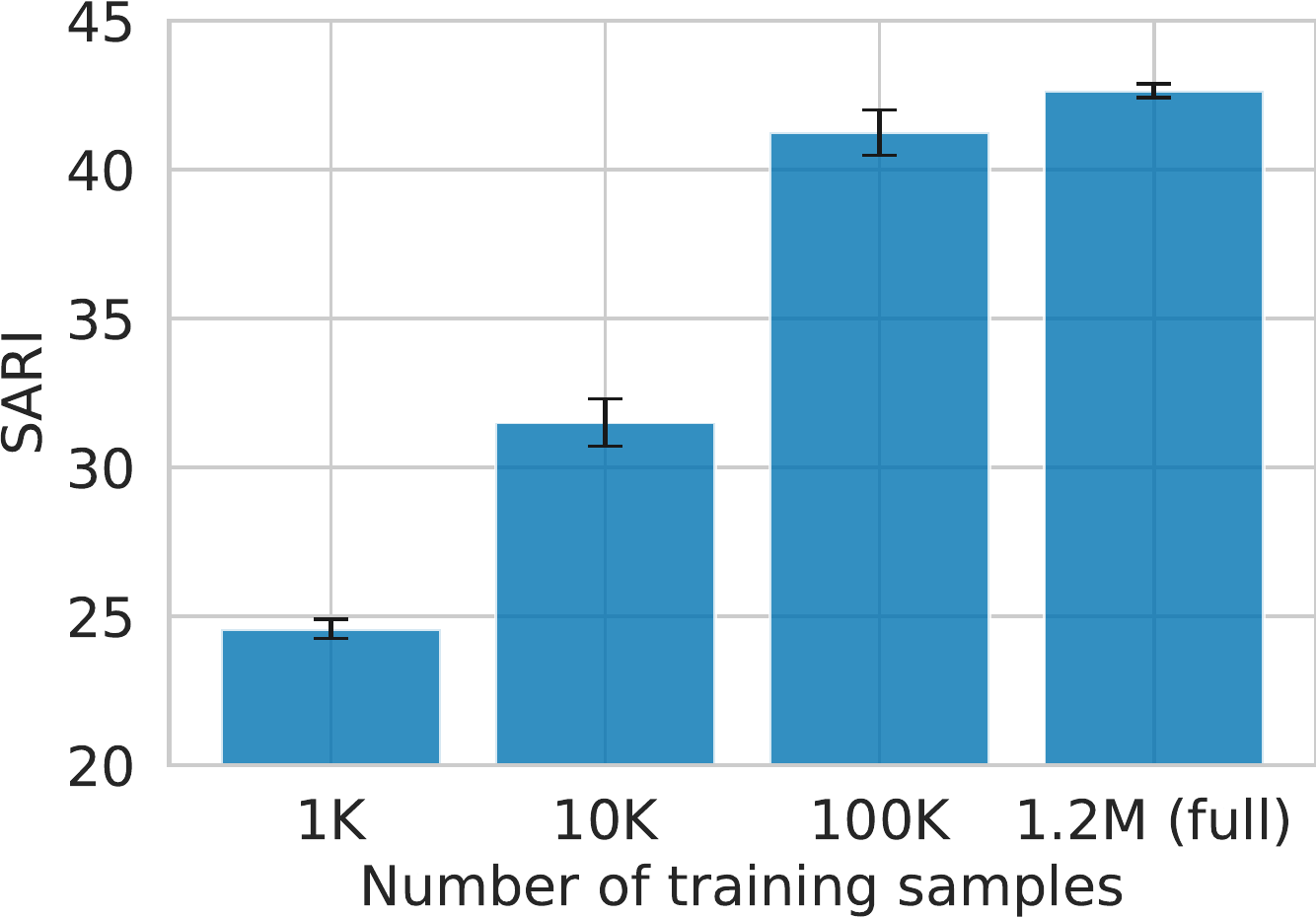}
    \caption{\label{subfigure:ablation_data_size} \textbf{Large-Scale Mining}}
\end{subfigure}
\hspace{1em}
\begin{subfigure}{0.30\textwidth}
    \centering
    \includegraphics[width=1\linewidth]{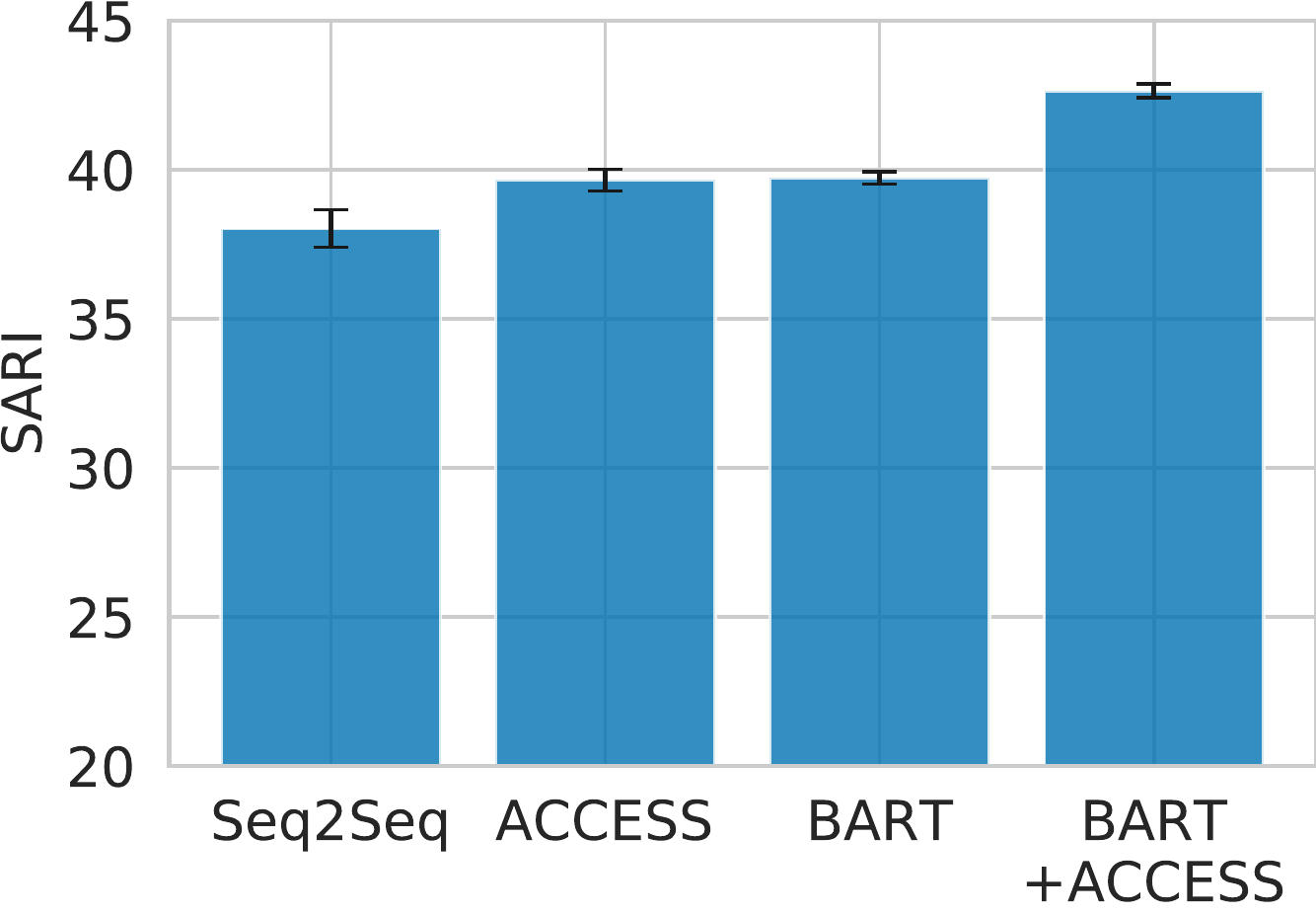}
    \caption{\label{subfigure:ablation_bart_and_access}\textbf{\bart and \access}}
\end{subfigure}
\caption{\textbf{Ablations} We display averaged SARI scores on the English \asset test set with 95\% confidence intervals (5 runs). (\subref{subfigure:ablation_simplifications_vs_paraphrases}) Models trained on mined simplifications or mined paraphrases, (\subref{subfigure:ablation_data_size}) \muss trained on varying amounts of mined data, (\subref{subfigure:ablation_bart_and_access}) Models trained with or without \bart and/or \access.}
\end{figure*}

\paragraph{Results}
\muss outperforms our strongest baseline by +8.25 SARI for French, while matching the pivot baseline performance for Spanish.

Besides using state-of-the-art machine translation models, the pivot baseline relies on a strong backbone simplification model that has two advantages compared to the French and Spanish simplification model.
First the simplification model of the pivot baseline was trained on labeled simplification data from \wikilarge, which obtains +1.5 SARI in English compared to training only on mined data.
Second it uses the stronger monolingual \bart model instead of \mbart. In Appendix Table~\ref{table:full_english_results}, we can see that \mbart has a small loss in performance of 1.54 SARI compared to its monolingual counterpart \bart, due to the fact that it handles 25 languages instead of one. Further improvements could be achieved by using monolingual \bart models trained for French or Spanish, possibly outperforming the pivot baseline.

\subsection{Human Evaluation}

\begin{table*}
\centering\small
\resizebox{\textwidth}{!}{
\begin{tabular}{l|r@{}l@{~}r@{}l@{~}r@{}l|r@{}l@{~}r@{}l@{~}r@{}l|r@{}l@{~}r@{}l@{~}r@{}l}
\toprule
& \multicolumn{6}{c|}{\textbf{English}} & \multicolumn{6}{c|}{\textbf{French}} & \multicolumn{6}{c}{\textbf{Spanish}} \\
 & \multicolumn{2}{c}{Adequacy} & \multicolumn{2}{c}{Fluency} & \multicolumn{2}{c|}{Simplicity} & \multicolumn{2}{c}{Adequacy} & \multicolumn{2}{c}{Fluency} & \multicolumn{2}{c|}{Simplicity} & \multicolumn{2}{c}{Adequacy} & \multicolumn{2}{c}{Fluency} & \multicolumn{2}{c}{Simplicity}\\ \midrule
\access \cite{martin2020controllable}  & \tnumem{3.10}{0.32} & \tnumem{3.46}{0.28} & \tnumem{1.40}{0.29} & ---& & ---&  & ---&   & ---&  & ---&  & ---&  \\
Pivot baseline  & ---&  & ---&  & ---&  & \tnumem{1.78}{0.40} & \tnumem{2.10}{0.47} & \tnumem{1.16}{0.31} & \tnumem{2.02}{0.28} & \tnumem{\mathbf{3.48}}{0.22} & \tnumem{2.20}{0.29} \\
Gold Reference  & \tnumem{\mathbf{3.44}}{0.28} & \tnumem{3.78}{0.17} & \tnumem{1.80}{0.28} & \tnumem{\mathbf{3.46}}{0.25} & \tnumem{\mathbf{3.92}}{0.10} & \tnumem{\mathbf{1.66}}{0.31} & \tnumem{2.18\dagger}{0.43} & \tnumem{3.38}{0.24} & \tnumem{1.26\dagger}{0.36} \\
\muss (\iffalse\mined\else unsup.\fi) & \tnumem{3.20}{0.28} & \tnumem{3.84}{0.14} & \tnumem{1.88}{0.33}  & \tnumem{2.88}{0.34} & \tnumem{3.50}{0.32} & \tnumem{1.22}{0.25} & \tnumem{\mathbf{2.26}}{0.29} & \tnumem{\mathbf{3.48}}{0.25} & \tnumem{\textbf{2.56}}{0.29} \\
\muss (\iffalse\textsc{Wiki.}+\mined\else sup.\fi) &  \tnumem{3.12}{0.34} & \tnumem{\mathbf{3.90}}{0.14} & \tnumem{\mathbf{2.22}}{0.36} &  ---& & ---&  & ---&   & ---&  & ---&  & ---& \\
\bottomrule
\end{tabular}
}
\caption{\label{table:human_eval} \textbf{Human Evaluation} We display human ratings of adequacy, fluency and simplicity for previous work \access,  pivot baseline, reference human simplifications, our best unsupervised systems (\bartaccess for English, \mbartaccess for other languages), and our best supervised model for English. Scores are averaged over 50 ratings per system with 95\% confidence intervals. $\dagger$Low ratings of the gold reference in Spanish \newsela is due to automatic alignment errors.}
\end{table*}

To further validate the quality of our models, we conduct a human evaluation in all languages according to adequacy, fluency, and simplicity and report the results in Table~\ref{table:human_eval}.

\paragraph{Human Ratings Collection} For human evaluation, we recruit volunteer native speakers for each language (5 in English, 2 in French, and 2 in Spanish).
We evaluate three linguistic aspects on a 5 point Likert scale (0-4): adequacy (\emph{is the meaning preserved?}), fluency (\emph{is the simplification fluent?}) and simplicity (\emph{is the simplification actually simpler?}).
For each system and each language, 50 simplifications are annotated and each simplification is rated once only by a single annotator.
The simplifications are taken from \asset (English), \alector (French), and \newsela (Spanish).

\paragraph{Discussion}
Table~\ref{table:human_eval} displays the average ratings along with 95\% confidence intervals. Human judgments confirm that our unsupervised and supervised \muss models are more fluent and produce simpler outputs than previous state-of-the-art  \cite{martin2019camembert}. They are deemed as fluent and simpler than the human simplifications from \asset test set, which indicates our model is able to reach a high level of simplicity thanks to the control mechanism.
In French and Spanish, our unsupervised model performs better or similar in all aspects than the strong supervised pivot baseline which has been trained on labeled English simplifications.
In Spanish the gold reference surprisingly obtains poor human ratings, which we found to be caused by errors in the automatic alignments of source sentences with simplified sentences of the same article, as previously highlighted in Subsection~\ref{subsection:evaluation_data}.

\subsection{Ablations}
\label{subsection:ablations}

\paragraph{Mining Simplifications vs.~Paraphrases}

In this work, we mined paraphrases to train simplification models. This has the advantage of making fewer assumptions earlier on, by keeping the mining and models as general as possible, so that they are able to adapt to more simplification scenarios.

We also compared to directly mining simplifications using simplification heuristics to make sure that the target side is simpler than the source, following previous work \cite{kajiwara-komachi-2016-building,surya2018unsupervised}.
To mine a simplification dataset, we followed the same paraphrase mining procedure of querying 1 billion sequences on an index of 1 billion sequences. Out of the resulting paraphrases, we kept only pairs that either contained sentence splits, reduced sequence length, or simpler vocabulary (similar to how previous work enforce an FKGL difference). We removed the paraphrase constraint that enforced sentences to be different enough. We tuned these heuristics to optimize SARI on the validation set. The resulting dataset has 2.7 million simplification pairs.
In Figure~\ref{subfigure:ablation_simplifications_vs_paraphrases}, we show that seq2seq models trained on mined paraphrases achieve better performance. A similar trend exists with \bart and \access, thus confirming that mining paraphrases can obtain better performance than mining simplifications.

\paragraph{How Much Mined Data Do You Need?} 

We investigate the importance of a scalable mining approach that can create million-sized training corpora for sentence simplification. In Figure~\ref{subfigure:ablation_data_size}, we analyze the performance of training our best model on English on different amounts of mined data. By increasing the number of mined pairs, SARI drastically improves, indicating that efficient mining at scale is critical to performance. Unlike human-created training sets, unsupervised mining allows for large datasets in multiple languages.

\paragraph{Improvements from Pretraining and Control}

We compare the respective influence of pretraining \bart and controllable generation \access in Figure~\ref{subfigure:ablation_bart_and_access}. While both \bart and \access bring improvement over standard sequence-to-sequence, they work best in combination.
Unlike previous approaches to text simplification, we use pretraining to train our simplification systems. We find that the main qualitative improvement from pretraining is increased fluency and meaning preservation. For example, in Appendix Table~\ref{table:examples_bart}, the model trained only with \access substituted \textit{culturally akin} with \textit{culturally much like}, but when using BART, it is simplified to the more fluent \textit{closely related}. 
While models trained on mined data see several million sentences, pretraining methods are typically trained on billions. Combining pretraining with controllable simplification enhances simplification performance by flexibly adjusting the type of simplification.
\section{Conclusion}

We propose a sentence simplification approach that does not rely on labeled  parallel simplification data thanks to controllable generation, pretraining and large-scale mining of paraphrases from the web.
This approach is language-agnostic and matches or outperforms previous state-of-the-art results, even from supervised systems that use labeled  simplification data, on three languages: English, French, and Spanish. 
In future work, we plan to investigate how to scale this approach to more languages and types of simplification, and to apply this method to paraphrase generation.
Another interesting direction for future work would to examine and improve factual consistency, especially related to named entity hallucination or disappearance.

\bibliography{bibliography}
\bibliographystyle{acl_natbib}

\clearpage
\appendix

\section*{Appendices}

\section{Experimental details} \label{sec:experimental_details}

\begin{figure}[!htbp]
     \centering
 \includegraphics[width=\columnwidth]{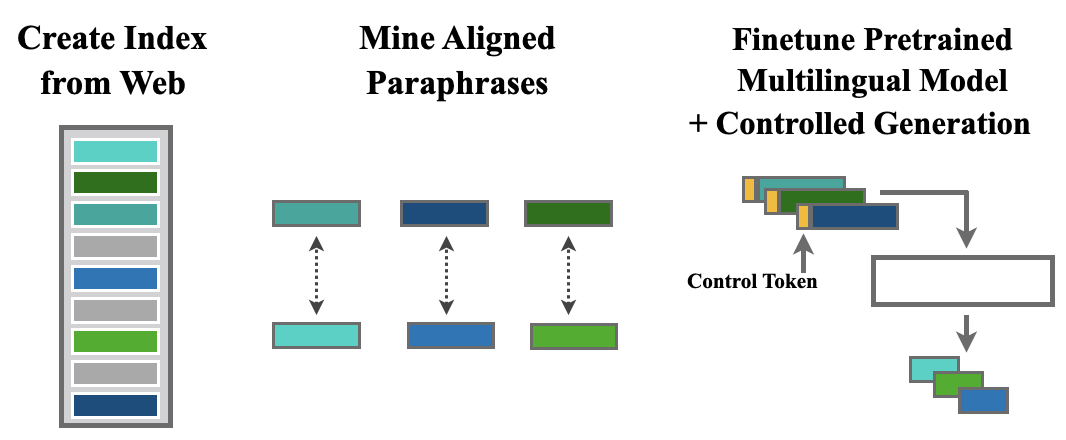}
     \caption{\textbf{Sentence Simplification Models for Any Language without Simplification Data}. Sentences from the web are used to create a large scale index that allows mining millions of paraphrases. Subsequently, we finetune pretrained models augmented with controllable mechanisms on the paraphrase corpora to achieve sentence simplification models in any language.}
     \label{fig:pullfig}
 \end{figure}

In this section we describe specific details of our experimental procedure. Figure~\ref{fig:pullfig} is a overall reminder of our method presented in the main paper.

\subsection{Mining Details} \label{sec:mining_details}

 \paragraph{Sequence Extraction} We only consider documents from the \textsc{head} split in \ccnet --- this represents the third of the data with the best perplexity using a  language model.

 \paragraph{Paraphrase Mining}
 We compute \laser embeddings of dimension 1024 and reduce  dimensionality with a 512 PCA followed by random rotation. We further compress them using 8 bit scalar quantization. The compressed embeddings are then stored in a \texttt{faiss} inverted file index with 32,768 cells (nprobe=16). These embeddings are used to mine pairs of paraphrases.
 We return the top-8 nearest neighbors, and keep those with L2 distance lower than 0.05 and relative distance compared to other top-8 nearest neighbors lower than 0.6.
 
 \paragraph{Paraphrases Filtering}
 The resulting paraphrases are filtered to remove almost identical paraphrases by enforcing a case-insensitive character-level Levenshtein distance \cite{levenshtein1966binary} greater or equal to 20\%. We remove paraphrases that come from the same document to avoid aligning sequences that overlapped each other in the  text. We also remove paraphrases where one of the sequence is contained in the other.
 We further filter out any sequence that is present in our evaluation datasets.

\subsection{Training Details}\label{subsection:training_details}
\paragraph{Seq2Seq training}
We implement our models with \texttt{fairseq}~\cite{ott2019fairseq}.
All our models are Transformers \cite{vaswani2017attention} based on the \bart\textsubscript{Large} architecture (388M parameters), keeping the optimization procedure and hyper-parameters fixed to those used in the original implementation \cite{lewis2019bart}\footnote{All hyper-parameters and training commands for \texttt{fairseq} can be found here: \url{https://github.com/pytorch/fairseq/blob/master/examples/bart/README.summarization.md}}.
We either randomly initialize weights for the standard sequence-to-sequence experiments or initialize with pretrained \bart for the \bart experiments. When initializing the weights randomly, we use a learning rate of $3.10^{-4}$ versus the original $3.10^{-5}$ when finetuning \bart. 
For a given seed, the model is trained on 8 Nvidia V100 GPUs during approximately 10 hours.

\paragraph{Controllable Generation}
For controllable generation, we use the open-source \access implementation \cite{martin2019reference}. We use the same control parameters as the original paper, namely length, Levenshtein similarity, lexical complexity, and syntactic complexity.\footnote{We modify the Levenshtein similarity parameter to only consider replace operations, by assigning a 0 weight to insertions and deletions. This change helps decorrelate the Levenshtein similarity control token from the length control token and produced better results in preliminary experiments.}

As mentioned in Section ''Simplifying with \access'', we select the 4 \access hyper-parameters using SARI on the validation set.
We use zero-order optimization with the \textsc{nevergrad} library \cite{nevergrad}. We use the OnePlusOne optimizer with a budget of 64 evaluations (approximately 1 hour of optimization on a single GPU).
The hyper-parameters are contained in the $[0.2, 1.5]$ interval.

The 4 hyper-parameter values are then kept fixed for all sentences in the associated test set.

\paragraph{Translation Model for Pivot Baseline}
For the pivot baseline we train models on \texttt{ccMatrix} \cite{schwenk2019ccmatrix}. Our models use the Transformer architecture with 240 million parameters with LayerDrop \cite{fan2019reducing}.
We train for 36 hours on 8 GPUs following the suggested parameters in \citet{ott2019fairseq}.

\paragraph{Gold Reference Baseline}
To avoid creating a discrepancy in terms of number of references between the gold reference scores, where we leave one reference out, and when we evaluate the models with all references, we compensate by duplicating one of the other references at random so that the total number of references is unchanged.

\subsection{Evaluation Details} \label{subsection:evaluation_details}

\paragraph{SARI score computation}
We use the latest version of SARI implemented in \texttt{EASSE} \cite{alva2019easse} which fixes bugs and inconsistencies from the traditional implementation of SARI. As a consequence, we also recompute scores from previous systems that we compare to. We do so by using the system predictions provided by the respective authors, and available in \texttt{EASSE}.

\paragraph{\alector Sentence-level Alignment}
The \alector corpus comes as source documents and their manual simplifications but not sentence-level alignment is provided.
Luckily, most of these documents were simplified line by line, each line consisting of a few sentences.
For each source document, we therefore align each line, provided it is not too long (less than 6 sentences), with the most appropriate line in the simplified document, using the \laser embedding space.
The resulting alignments are split into validation and test by randomly sampling the documents for the validation (450 sentence pairs) and rest for test (416 sentence pairs). 

\subsection{Links to Datasets}
The datasets we used are available at the following addresses:
\begin{itemize}
    \item \ccnet: \\\url{https://github.com/facebookresearch/cc_net}.
    \item \wikilarge: \\\url{https://github.com/XingxingZhang/dress}.
    \item \asset: \\\url{https://github.com/facebookresearch/asset} or \url{https://github.com/feralvam/easse}.
    \item \turkcorpus: \\\url{https://github.com/cocoxu/simplification/} or \url{https://github.com/feralvam/easse}.
    \item \newsela: This dataset has to be requested at \url{https://newsela.com/data}.
    \item \alector: This dataset has to be requested from the authors \cite{gala2020alector}.
\end{itemize}

\section{Characteristics of the mined data}
\begin{figure*}
    \centering
    \includegraphics[width=\linewidth]{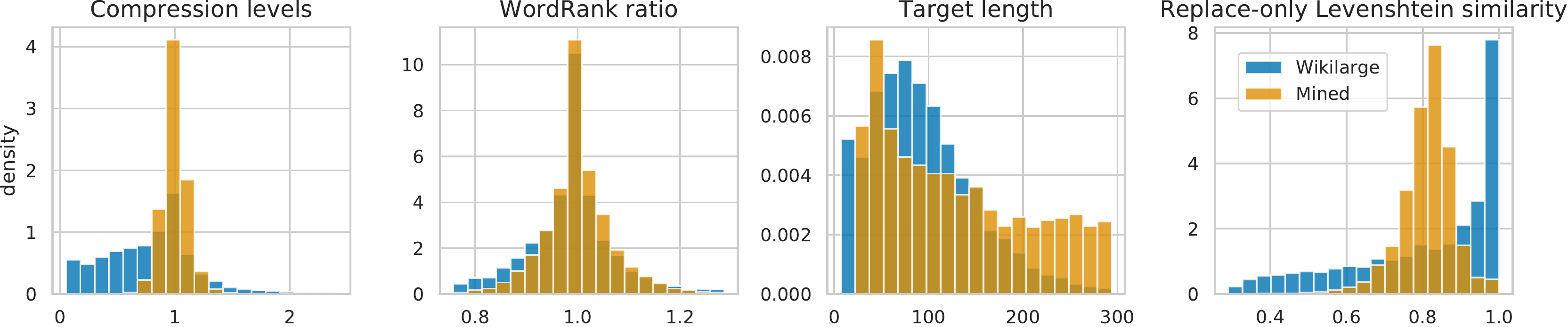}
    \caption{Density of several text features in \wikilarge and our mined data. The WordRank ratio is a measure of lexical complexity reduction \cite{martin2020controllable}. Replace-only Levenshtein similarity only considers replace operations in the traditional Levenshtein similarity and assigns 0 weights to insertions and deletions.}
    \label{figure:dataset_features}
\end{figure*}

We show in Figure~\ref{figure:dataset_features} the distribution of different surface features of our mined data versus those of \wikilarge. Some examples of mined paraphrases are shown in Table~\ref{table:mining_examples}.

\begin{table*}[t]
\centering
\small
\begin{tabular}{@{}lp{0.87\textwidth}@{}}
\toprule
\textbf{Query} & \textbf{For} \textbf{insulation,} \textbf{it} \textbf{uses} \textbf{foam-injected} polyurethane which helps \textbf{ensure} the quality of the ice \textbf{produced} \textbf{by} \textbf{the} \textbf{machine.} \textbf{It} \textbf{comes} \textbf{with} an easy to clean air filter. \\
\textbf{Mined} & \textbf{It} \textbf{has} polyurethane \textbf{for} \textbf{insulation} which \textbf{is} \textbf{foam-injected.} \textbf{This} helps \textbf{to} \textbf{maintain} the quality of the ice \textbf{it} \textbf{produces.} \textbf{The} \textbf{unit} \textbf{has} an easy to clean air filter. \\
\midrule

\textbf{Query} & Here are some \textbf{useful} tips and \textbf{tricks} to \textbf{identify} \textbf{and} manage your \textbf{stress.} \\
\textbf{Mined} & Here are some tips and \textbf{remedies} \textbf{you} \textbf{can} \textbf{follow} to manage \textbf{and} \textbf{control} your \textbf{anxiety.} \\
\midrule

\textbf{Query} & \textbf{As} \textbf{cancer} cells \textbf {break} \textbf{apart,} their contents are \textbf{released} into the \textbf{blood.} \\
\textbf{Mined} & \textbf{When} \textbf{brain} cells \textbf{die,} their contents are \textbf{partially} \textbf{spilled} \textbf{back} into the \textbf{blood} \textbf{in} \textbf{the} \textbf{form} \textbf{of} \textbf{debris.} \\
\midrule

\textbf{Query} & \textbf{The} \textbf{trail} is ideal for \textbf{taking} a short \textbf{hike} \textbf{with} \textbf{small} \textbf{children} or a \textbf{longer,} \textbf{more} \textbf{rugged} \textbf{overnight} \textbf{trip.} \\
\textbf{Mined} & \textbf{It} is \textbf{the} ideal \textbf{location} for a short \textbf{stroll,} \textbf{a} \textbf{nature} \textbf{walk} or a \textbf{longer} \textbf{walk.} \\
\midrule
\textbf{Query} & Thank you for \textbf{joining} \textbf{us,} \textbf{and} \textbf{please} check \textbf{out} the \textbf{site.} \\
\textbf{Mined} & Thank you for \textbf{calling} \textbf{us.} \textbf{Please} check the \textbf{website.} \\
\midrule
\end{tabular}
\caption{\textbf{Examples of Mined Paraphrases}. Paraphrases, although sometimes not preserving the entire meaning, display various rewriting operations, such as lexical substitution, compression or sentence splitting.}
\label{table:mining_examples}
\end{table*}

\section{Set \access Control Parameters Without Parallel Data} \label{section:prior_knowledge_values}

\begin{table}[!htbp]
\centering
\small
\resizebox{\columnwidth}{!}{
\begin{tabular}{l|l|l|l}
\toprule
 & \textbf{\asset} & \textbf{\turkcorpusabbr} & \textbf{\newsela} \\
 Method & SARI $\uparrow$ & SARI $\uparrow$ & SARI $\uparrow$\\
\midrule
 SARI on valid & \numem{42.65}{0.23} & \numem{40.85}{0.15} & \numem{38.09}{0.59} \\
 Approx. value & \numem{42.49}{0.34} & \numem{39.57}{0.40} & \numem{36.16}{0.35} \\
\bottomrule
\end{tabular}
}
\caption{\textbf{Set \access Controls Wo. Parallel Data}\\
Setting \access parameters of \muss+\mined model either using SARI on the validation set or using only 50 \textit{unaligned} sentence pairs from the validation set. All \access parameters are set to the same approximated value: \asset$=0.8$, \turkcorpus$=0.95$, and \newsela$=0.4$).
\label{table:prior_knowledge_values}}
\end{table}

In our experiments we adjusted our model to the different dataset conditions by selecting our \access control tokens with SARI on each validation set.
When no such parallel validation set exists, we show that strong performance can still be obtained by using prior knowledge for the given downstream application. This can be done by setting all 4 \access control hyper-parameters to an intuitive guess of the desired compression ratio.

To illustrate this for the considered evaluation datasets, we first independently sample 50 source sentences and 50 random \textit{unaligned} simple sentences from each validation set. These two groups of non-parallel sentences are used to approximate the character-level compression ratio between complex and simplified sentences.
We do so by dividing the average length of the simplified sentences by the average length of the 50 source sentences.
We finally use this approximated compression ratio as the value of all 4 \access hyper-parameters.
In practice, we obtain the following approximations: \asset$=0.8$, \turkcorpus$=0.95$, and \newsela$=0.4$ (rounded to $0.05$).
Results in Table~\ref{table:prior_knowledge_values} show that the resulting model performs very close to when we adjust the \access hyper-parameters using SARI on the complete validation set.

\section{Comparing to Existing Paraphrase Datasets}\label{section:comparison_with_paraphrase_dataset}
We compare using our mined paraphrase data with existing large-scale paraphrase datasets in Table~\ref{table:comparison_with_paraphrase_dataset}.
We use \textsc{ParaNMT} \cite{wieting-gimpel-2018-paranmt}, a large paraphrase dataset created using back-translation on an existing labeled parallel machine translation dataset.
We use the same 5 million top-scoring sentences that the authors used to train their sentence embeddings.
Training \muss on the mined data or on \textsc{ParaNMT} obtains similar results for text simplification, confirming that mining paraphrase data is a viable alternative to using existing paraphrase datasets relying on labeled parallel machine translation corpora.

\begin{table}[!htbp]
\centering
\small
\resizebox{\columnwidth}{!}{
\begin{tabular}{l|l|l|l}
\toprule
 & \textbf{\asset} & \textbf{\turkcorpusabbr} & \textbf{\newsela} \\
 Data & SARI $\uparrow$ & SARI $\uparrow$ & SARI $\uparrow$\\
\midrule
\mined & \numem{42.65}{0.23} & \numem{40.85}{0.15} & \numem{38.09}{0.59} \\
\textsc{ParaNMT} & \numem{42.50}{0.33} & \numem{40.50}{0.16} & \numem{39.11}{0.88} \\
\bottomrule
\end{tabular}
}
\caption{\textbf{Mined Data vs. ParaNMT}\\
We compare SARI scores of \muss trained either on our mined data or on \textsc{ParaNMT} \cite{wieting-gimpel-2018-paranmt} on the test sets of \asset, \turkcorpus and \newsela.
\label{table:comparison_with_paraphrase_dataset}}
\end{table}

\section{Influence of BART on Fluency}
In Table~\ref{table:examples_bart}, we present some selected samples that highlight the improved fluency of simplifications when using \bart.
\begin{table*}
\centering
\small
\begin{tabular}{@{}lp{0.55\textwidth}@{}}
\toprule
 \textbf{Original} & They are culturally akin to the coastal peoples of Papua New Guinea. \\
 \textbf{\access} & \textbf{They're} culturally \textbf{much like} the Papua New \textbf{Guinea coastal peoples.} \\
 \textbf{\bartaccess} & They are \textbf{closely related} to coastal \textbf{people} of Papua New Guinea \\
 \midrule
\textbf{Original} & Orton and his wife welcomed Alanna Marie Orton on July 12, 2008. \\
\textbf{\access} & Orton and his wife \textbf{had been called} Alanna Marie Orton on July \textbf{12}. \\
 \textbf{\bartaccess} & Orton and his wife \textbf{gave birth to} Alanna Marie Orton on July 12, 2008. \\
 \midrule
\textbf{Original} & He settled in London, devoting himself chiefly to practical teaching. \\
\textbf{\access} & He \textbf{set up} in \textbf{London and made} himself \textbf{mainly for} teaching. \\
 \textbf{\bartaccess} & He settled in \textbf{London and devoted} himself to teaching. \\
\bottomrule
\end{tabular}
\caption{\textbf{Influence of \bart on Simplifications.} We display some examples of generations that illustrate how \bart improves the fluency and meaning preservation of generated simplifications.\label{table:examples_bart}}

\end{table*}

\section{Additional Scores}
\paragraph{BLEU} We report additional BLEU scores for completeness. These results are displayed along with SARI and FKGL for English. %
These BLEU scores should be carefully interpreted. They have been found to correlate poorly with human judgments of simplicity \cite{sulem2018semantic}. Furthermore, the identity baseline achieves very high BLEU scores on some datasets (e.g. 92.81 on \asset or 99.36 on \turkcorpus), which underlines the weaknesses of this metric.

\paragraph{Validation Scores} We report English validation scores %
to foster reproducibility in Table~\ref{table:full_english_results_valid}.%

\paragraph{Seq2Seq Models on Mined Data}
When training a Transformer sequence-to-sequence  model (Seq2Seq) on \wikilarge compared to the mined corpus, models trained on the mined data perform better. %
It is surprising that a model trained solely on paraphrases achieves such good results on simplification benchmarks.
Previous works have shown that simplification models suffer from not making enough modifications to the source sentence and found that forcing models to rewrite the input was beneficial \cite{wubben2012sentence,martin2020controllable}. This is confirmed when investigating the F1 deletion component of SARI which is 20 points higher for the model trained on paraphrases.

\begin{table*}
\centering\small
\resizebox{\textwidth}{!}{
\begin{tabular}{ll|lll|lll|lll}
\toprule
 & \textbf{Data} & \multicolumn{3}{c|}{\textbf{\asset}} & \multicolumn{3}{c|}{\textbf{\turkcorpus}} & \multicolumn{3}{c}{\textbf{\newsela}} \\

\multicolumn{2}{l|}{\textbf{\textit{Baselines and Gold Reference}}} & SARI $\uparrow$ & BLEU $\uparrow$ &  FKGL $\downarrow$ & SARI $\uparrow$ & BLEU $\uparrow$ &  FKGL $\downarrow$ & SARI $\uparrow$ & BLEU $\uparrow$ &  FKGL $\downarrow$ \\
\midrule
Identity Baseline & --- & 20.73 & 92.81 & 10.02 & 26.29 & 99.36 & 10.02 & --- & --- & --- \\
Truncate Baseline & --- & 29.85 & 84.94 & 7.91 & 33.10 & 88.82 & 7.91 & --- & --- & --- \\
Reference & --- & \numem{44.87}{0.36} & \numem{68.95}{1.33} & \numem{6.49}{0.15} & \numem{40.04}{0.30} & \numem{73.56}{1.18} & \numem{8.77}{0.08} & --- & --- & --- \\

\midrule
\\[-2mm]
\multicolumn{2}{l}{\textbf{\textit{Supervised Systems (This Work)}}} \\
\midrule

Seq2Seq & \wikilarge & \numem{32.71}{1.55} & \numem{88.56}{1.06} & \numem{8.62}{0.34} & \numem{35.79}{0.89} & \numem{90.24}{2.52} & \numem{8.63}{0.34} & \numem{22.23}{1.99} & \numem{21.75}{0.45} & \numem{8.00}{0.26} \\
\muss & \wikilarge & \numem{43.63}{0.71} & \numem{76.28}{4.30} & \numem{6.25}{0.42} & \numem{42.62}{0.27} & \numem{78.28}{3.95} & \numem{6.98}{0.95} & \numem{40.00}{0.63} & \numem{14.42}{6.85} & \numem{3.51}{0.53} \\
\muss & \wikilarge + \mined & \numem{44.15}{0.56} & \numem{72.98}{4.27} & \numem{6.05}{0.51} & \numem{42.53}{0.36} & \numem{78.17}{2.20} & \numem{7.60}{1.06} & \numem{39.50}{0.42} & \numem{15.52}{0.99} & \numem{3.19}{0.49} \\
\muss & \newsela & \numem{42.91}{0.58} & \numem{71.40}{6.38} & \numem{6.91}{0.42} & \numem{41.53}{0.36} & \numem{74.29}{4.67} & \numem{7.39}{0.42} & \numem{42.59}{1.00} & \numem{18.61}{4.49} & \numem{2.74}{0.98} \\
\muss & \newsela + \mined & \numem{41.36}{0.48} & \numem{78.35}{2.83} & \numem{6.96}{0.26} & \numem{40.01}{0.51} & \numem{83.77}{1.00} & \numem{8.26}{0.36} & \numem{41.17}{0.95} & \numem{16.87}{4.55} & \numem{2.70}{1.00} \\

\midrule
\\[-2mm]
\multicolumn{2}{l}{\textbf{\textit{Unsupervised Systems (This Work)}}} \\
\midrule

Seq2Seq & \mined & \numem{38.03}{0.63} & \numem{61.76}{2.19} & \numem{9.41}{0.07} & \numem{38.06}{0.47} & \numem{63.70}{2.43} & \numem{9.43}{0.07} & \numem{30.36}{0.71} & \numem{12.98}{0.32} & \numem{8.85}{0.13} \\
\muss (\mbart) & \mined & \numem{41.11}{0.70} & \numem{77.22}{2.12} & \numem{7.18}{0.21} & \numem{39.40}{0.54} & \numem{77.05}{3.02} & \numem{8.65}{0.40} & \numem{34.76}{0.96} & \numem{19.06}{1.15} & \numem{5.44}{0.25} \\
\muss (\bart) & \mined & \numem{42.65}{0.23} & \numem{66.23}{4.31} & \numem{8.23}{0.62} & \numem{40.85}{0.15} & \numem{63.76}{4.26} & \numem{8.79}{0.30} & \numem{38.09}{0.59} & \numem{14.91}{1.39} & \numem{5.12}{0.47} \\

\bottomrule
\end{tabular}

}
\caption{\label{table:full_english_results}
\textbf{Detailed English Results.} We display SARI, BLEU, and FKGL on \asset, \turkcorpus and \newsela English evaluation datasets (test sets).\\
}
\end{table*}

\begin{table*}
\centering\small
\resizebox{\textwidth}{!}{
\begin{tabular}{ll|lll|lll|lll}
\toprule
 & \textbf{Data} & \multicolumn{3}{c|}{\textbf{\asset}} & \multicolumn{3}{c|}{\textbf{\turkcorpus}} & \multicolumn{3}{c}{\textbf{\newsela}} \\

\multicolumn{2}{l|}{\textbf{\textit{Baselines and Gold Reference}}} & SARI $\uparrow$ & BLEU $\uparrow$ &  FKGL $\downarrow$ & SARI $\uparrow$ & BLEU $\uparrow$ &  FKGL $\downarrow$ & SARI $\uparrow$ & BLEU $\uparrow$ &  FKGL $\downarrow$ \\
\midrule

Identity Baseline & --- & 22.53 & 94.44 & 9.49 & 26.96 & 99.27 & 9.49 & 12.00 & 20.69 & 8.77 \\
Truncate Baseline & --- & 29.95 & 86.67 & 7.39 & 32.90 & 89.10 & 7.40 & 24.64 & 18.97 & 6.90 \\
Reference & --- & \numem{45.22}{0.94} & \numem{72.67}{2.83} & \numem{6.13}{0.56} & \numem{40.66}{0.11} & \numem{77.21}{0.45} & \numem{8.31}{0.04} & --- & --- & --- \\

\midrule
\\[-2mm]
\multicolumn{2}{l}{\textbf{\textit{Supervised Systems (This Work)}}} \\
\midrule

Seq2Seq & \wikilarge & \numem{33.87}{1.90} & \numem{90.21}{1.14} & \numem{8.31}{0.34} & \numem{35.87}{1.09} & \numem{91.06}{2.24} & \numem{8.31}{0.34} & \numem{20.89}{4.08} & \numem{20.97}{0.53} & \numem{8.27}{0.46} \\
\muss & \wikilarge & \numem{45.58}{0.28} & \numem{78.85}{4.44} & \numem{5.61}{0.31} & \numem{43.26}{0.42} & \numem{78.39}{3.08} & \numem{6.73}{0.38} & \numem{39.66}{1.80} & \numem{14.82}{7.17} & \numem{4.64}{1.85} \\
\muss & \wikilarge + \mined & \numem{45.50}{0.69} & \numem{73.16}{4.41} & \numem{5.83}{0.51} & \numem{43.17}{0.19} & \numem{77.52}{3.01} & \numem{7.19}{1.02} & \numem{40.50}{0.56} & \numem{16.30}{0.97} & \numem{3.57}{0.60} \\
\muss & \newsela & \numem{43.91}{0.10} & \numem{70.06}{10.05} & \numem{6.47}{0.29} & \numem{41.94}{0.21} & \numem{74.03}{7.51} & \numem{6.99}{0.53} & \numem{42.36}{1.32} & \numem{19.18}{6.03} & \numem{3.20}{1.01} \\
\muss & \newsela + \mined & \numem{42.48}{0.41} & \numem{77.86}{3.13} & \numem{6.41}{0.13} & \numem{40.77}{0.52} & \numem{83.04}{1.16} & \numem{7.68}{0.30} & \numem{41.68}{1.60} & \numem{17.23}{5.28} & \numem{2.97}{0.91} \\

\midrule
\\[-2mm]
\multicolumn{2}{l}{\textbf{\textit{Unsupervised Systems (This Work)}}} \\
\midrule
Seq2Seq & \mined & \numem{38.88}{0.22} & \numem{61.80}{0.94} & \numem{8.63}{0.13} & \numem{37.51}{0.10} & \numem{62.04}{0.91} & \numem{8.64}{0.13} & \numem{30.35}{0.23} & \numem{13.04}{0.45} & \numem{8.87}{0.12} \\
\muss (\mbart) & \mined & \numem{41.68}{0.72} & \numem{77.11}{2.02} & \numem{6.56}{0.21} & \numem{39.60}{0.44} & \numem{75.64}{2.85} & \numem{8.04}{0.40} & \numem{34.59}{0.59} & \numem{18.19}{1.26} & \numem{5.76}{0.22} \\
\muss (\bart) & \mined & \numem{43.01}{0.23} & \numem{67.65}{4.32} & \numem{7.75}{0.53} & \numem{40.61}{0.18} & \numem{63.56}{4.30} & \numem{8.28}{0.18} & \numem{38.07}{0.22} & \numem{14.43}{0.97} & \numem{5.40}{0.41} \\

\bottomrule
\end{tabular}

}
\caption{\label{table:full_english_results_valid} \textbf{English Results on Validation Sets.} We display SARI, BLEU, and FKGL on \asset, \turkcorpus and \newsela English datasets (validation sets).}
\end{table*}

\end{document}